\documentclass[10pt,twocolumn,letterpaper]{article}
\usepackage{iccv}
\usepackage{times}
\usepackage{epsfig}
\usepackage{graphicx}
\usepackage{amsmath}
\usepackage{amssymb}
\usepackage{subfigure}

\newcommand\ceil[1]{\lceil#1\rceil}
\usepackage[pagebackref=true,breaklinks=true,letterpaper=true,colorlinks,bookmarks=false]{hyperref}

\iccvfinalcopy 

\def\iccvPaperID{762} 

\ificcvfinal\fi
\begin{document}

\title{Learning Convolutional Networks for Content-weighted Image Compression}

\author{Mu Li\\
Hong Kong Polytechnic University\\
{\tt\small csmuli@comp.polyu.edu.hk}
\and
Wangmeng Zuo\thanks{Corresponding Author}\\
Harbin Institute of Technology\\
{\tt\small cswmzuo@gmail.com}
\and
Shuhang Gu\\
Hong Kong Polytechnic University\\
{\tt\small shuhanggu@gmail.com}
\and
Debin Zhao\\
Harbin Institute of Technology\\
{\tt\small dbzhao@hit.edu.cn}
\and
David Zhang\\
Hong Kong Polytechnic University\\
{\tt\small csdzhang@comp.polyu.edu.hk}
}

\maketitle

\begin{abstract}
Lossy image compression is generally formulated as a joint rate-distortion optimization to learn encoder, quantizer, and decoder.
However, the quantizer is non-differentiable, and discrete entropy estimation usually is required for rate control.
These make it very challenging to develop a convolutional network (CNN)-based image compression system.
In this paper, motivated by that the local information content is spatially variant in an image, we suggest that the bit rate of the different parts of the image should be adapted to local content.
And the content aware bit rate is allocated under the guidance of a content-weighted importance map.
Thus, the sum of the importance map can serve as a continuous alternative of discrete entropy estimation to control compression rate.
And binarizer is adopted to quantize the output of encoder due to the binarization scheme is also directly defined by the importance map.
Furthermore, a proxy function is introduced for binary operation in backward propagation to make it differentiable.
Therefore, the encoder, decoder, binarizer and importance map can be jointly optimized in an end-to-end manner by using a subset of the ImageNet database.
In low bit rate image compression, experiments show that our system significantly outperforms JPEG and JPEG 2000 by structural similarity (SSIM) index, and can produce the much better visual result with sharp edges, rich textures, and fewer artifacts.
\end{abstract}

\section{Introduction}
Image compression is a fundamental problem in computer vision and image processing.
With the development and popularity of high-quality multimedia content, lossy image compression has been becoming more and more essential in saving transmission bandwidth and hardware storage.
An image compression system usually includes three components, \ie encoder, quantizer, and decoder, to form the codec.
The typical image encoding standards, \eg, JPEG and JPEG 2000, generally rely on handcrafted image transformation and separate optimization on codecs, and thus are suboptimal for image compression.
Moreover, JPEG and JPEG 2000 perform poor for low rate image compression, and usually are inevitable in producing some visual artifacts, \eg, blurring, ringing, and blocking.

Recently, deep convolutional networks (CNNs) have achieved great success in versatile vision tasks \cite{krizhevsky2012imagenet,Parkhi15,girshick2014rich,
xie2012image,dong2014learning}.
As to image compression, CNN is also expected to be more powerful than JPEG and JPEG 2000 by considering the following reasons.
First, for image encoding and decoding, flexible nonlinear analysis and synthesis transformations can be easily deployed by stacking several convolutional layers.
Second, it allows to jointly optimize the nonlinear encoder and decoder in an end-to-end manner.
Furthermore, several recent advances also validate the effectiveness of deep learning in image compression \cite{toderici2015variable,toderici2016full,
balle2016end,theis2017lossy}.

However, there are still several issues to be addressed in CNN-based image compression.
In general, lossy image compression can be formulated as a joint rate-distortion optimization to learn encoder, quantizer, and decoder.
Even the encoder and decoder can be represented as CNNs and be optimized via back-propagation,
the learning of non-differentiable quantizer is still a challenge problem. 
Moreover, the system aims to jointly minimize both the compression rate and distortion, where entropy rate should also be estimated and minimized in learning.
As a result of quantization, the entropy rate defined on discrete codes is also a discrete function, and a continuous approximation is required.

\begin{figure*}
\begin{center}
\includegraphics[width=1\linewidth]{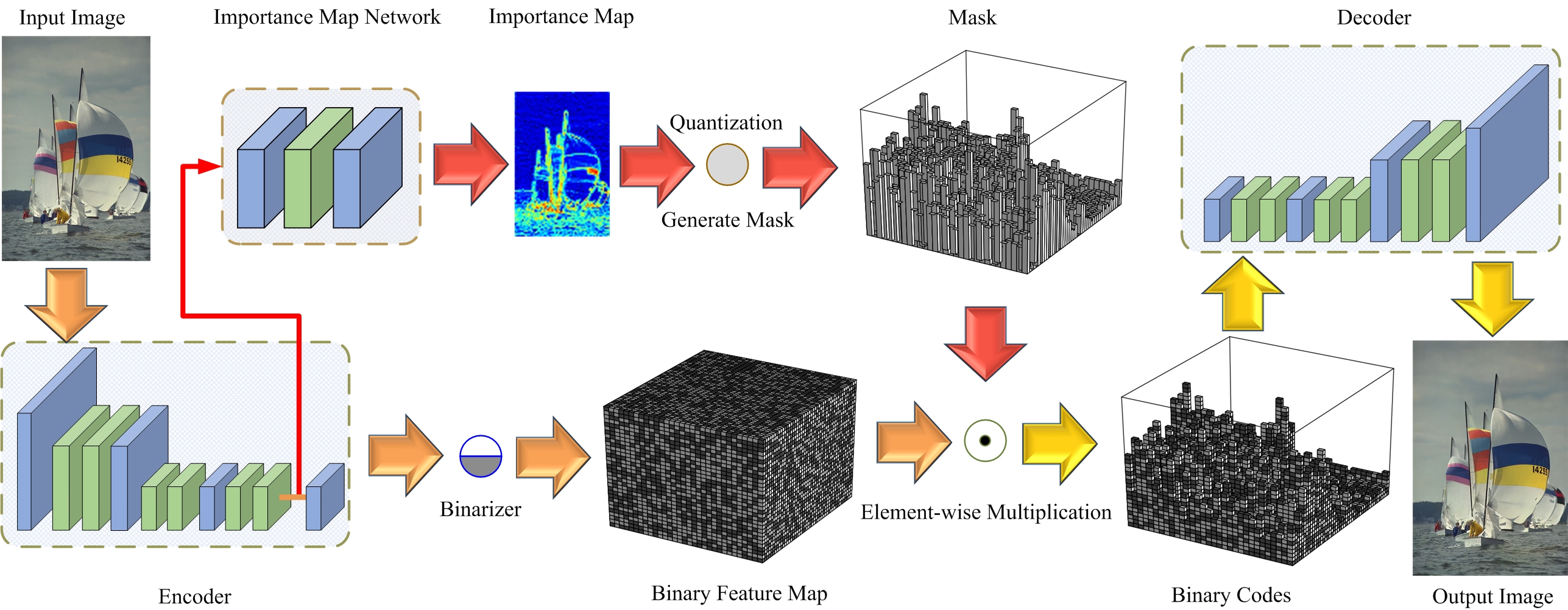}
\end{center}
   \caption{Illustration of the CNN architecture for content-weighted image compression.}
\label{frame_work}
\end{figure*}

In this paper, we present a novel CNN-based image compression framework to address the issues raised by quantization and entropy rate estimation.
For the existing deep learning based compression models~\cite{toderici2015variable,toderici2016full,balle2016end}, the discrete code after quantization should first have the same length with the encoder output, and then compressed based on entropy coding.
That is, the discrete code before entropy coding is spatially invariant.
However, it is generally known that the local information content is spatially variant in an image.
Thus, the bit rate should also be spatially variant to adapt to local information content.
To this end, we introduce a content-weighted importance map to guide the allocation of local bit rate.
Given an input image $\mathbf{x}$, let $\mathbf{e} = E(\mathbf{x})$ be the output of encoder network, which includes $n$ feature maps with size of $h \times w$.
Denote by $\mathbf{p} = P(\mathbf{x})$ the $h \times w$ non-negative importance map.
Specifically, when $\frac{l-1}{L} \leq \mathbf{p}_{i,j} < \frac{l}{L}$, we will only encode the first $\frac{nl}{L}$-{th} feature maps at spatial location $(i,j)$.
Here, $L$ is the number of the importance level. And $\frac{n}{L}$ is the number of bits for each importance level.
The other feature maps are automatically set with 0 and need not be saved into the codes.
By this way, we can allocate more bits to the region with rich content, which is very helpful in preserving texture details with less sacrifice of bit rate.
Moreover, the sum of the importance map $\sum_{i,j} \mathbf{p}_{i,j}$ will serve as a continuous estimation of compression rate, and can be directly adopted as a compression rate controller.

Benefited from importance map, we do not require to use any other entropy rate estimate in our objective, and can adopt a simple binarizer for quantization.
The binarizer set those features with the possibility over 0.5 to 1 and others to 0.
Inspired by the binary CNN~\cite{zhou2016dorefa,rastegari2016xnor,courbariaux2016binarized}, we introduce a proxy function for the binary operation in backward propagation and make it trainable.
As illustrated in Figure~\ref{frame_work}, our compression framework consists of four major components: convolutional encoder, importance map network, binarizer, and convolutional decoder.
With the introduction of continuous importance map and proxy function, all the components can be jointly optimized in an end-to-end manner.

Note that we do not include any term on entropy rate estimate in the training of the compression system.
And the local spatial context is also not utilized.
Therefore, we design a convolutional entropy coder to predict the current code with its context, and apply it to context-adaptive binary arithmetic coding (CABAC) framework~\cite{marpe2003context} to further compress the binary codes and importance map.


Our whole framework is trained on a subset of the ImageNet database and tested on the Kodak dataset.
In low bit rate image compression, our system achieves much better rate-distortion performance than JPEG and JPEG 2000 in terms of both quantitative metrics (\eg, SSIM and MSE) and visual quality.
More remarkably, the compressed images by our system are visually more pleasant in producing sharp edges, rich textures, and fewer artifacts.
Compared with other CNN-based system~\cite{toderici2015variable,toderici2016full,balle2016end}, ours performs better in retaining texture details while suppressing visual artifacts.
%

To sum up, the main contribution of this paper is to introduce content-weighted importance map and binary quantization in the image compression system.
The importance map not only can be used to substitute entropy rate estimate in joint rate-distortion optimization, but also can be adopted to guide the local bit rate allocation.
By equipping with binary quantization and the proxy function, our compression system can be end-to-end trained, and obtain significantly better results than JPEG and JPEG 2000. 


\section{Related Work}
For the existing image standards, \eg, JPEG and JPEG 2000, the codecs actually are separately optimized.
In the encoding stage, they first perform a linear transform to an image.
Quantization and lossless entropy coding are then utilized to minimize the compression rate.
%
%
For example, JPEG ~\cite{wallace1992jpeg} applies discrete cosine transform (DCT) on $8\times8$ image patches, quantizes the frequency components and compresses the quantized codes with a variant of Huffman encoding.
JPEG 2000 ~\cite{skodras2001jpeg} uses a multi-scale orthogonal wavelet decomposition to transform an image, and encodes the quantized codes with the Embedded Block Coding with Optimal Truncation.
In the decoding stage, decoding algorithm and inverse transform are designed to minimize distortion.
In contrast, we model image compression as a joint rate-distortion optimization, where both nonlinear encoder and decoder are jointly trained in an end-to-end manner.

%
%

Recently, several deep learning based image compression models have been developed.
For lossless image compression, deep learning models have achieved state-of-the-art performance~\cite{theis2015generative,oord2016pixel}.
For lossy image compression, Toderici \etal~\cite{toderici2015variable} present a recurrent neural network (RNN) to compress $32 \times 32$ images.
Toderici \etal~\cite{toderici2016full} further introduce a set of full-resolution compression methods for progressive encoding and decoding of images.
These methods learn the compression models by minimizing the distortion for a given compression rate.
While our model is end-to-end trained via joint rate-distortion optimization.

The most related work is that of~\cite{balle2016end,theis2017lossy} based on convolutional autoencoders.
Ball{\'e} \etal~\cite{balle2016end} use generalized divisive normalization (GDN) for joint nonlinearity, and replace rounding quantization with additive uniform noise for continuous relaxation of distortion and entropy rate loss.
Theis \etal~\cite{theis2017lossy} adopt a smooth approximation of the derivative of the rounding function, and upper-bound the discrete entropy rate loss for continuous relaxation.
%
%
Our content-weighted image compression system is different with~\cite{balle2016end,theis2017lossy} in rate loss, quantization, and continuous relaxation.
Instead of rounding and entropy, we define our rate loss on importance map and adopt a simple binarizer for quantization.
Moreover, the code length after quantization is spatially invariant in~\cite{balle2016end,theis2017lossy}.
In contrast, the local code length in our model is content-aware, which is useful in improving visual quality.


Our work is also related to binarized neural network (BNN)~\cite{courbariaux2016binarized}, where both weights and activations are binarized to $+1$ or $-1$ to save memory storage and run time.
Courbariaux \etal~\cite{courbariaux2016binarized} adopt a straight-through estimator to compute the gradient of the binarizer.
In our compression system, only the encoder output is binarized to $1$ or $0$, and a similar proxy function is used in backward propagation.



\section{Content-weighted Image Compression} \label{sec3}

Our content-weighted image compression framework is composed of four components, \ie convolutional encoder, binarizer, importance map network, and convolutional decoder.
And Figure~\ref{frame_work} shows the whole network architecture.
Given an input image $\mathbf{x}$, the convolutional encoder defines a nonlinear analysis transform by stacking convolutional layers, and outputs $E(\mathbf{x})$.
The binarizer $B(E(\mathbf{x}))$ assigns 1 to the encoder output higher than 0.5, and 0 to the others.
The importance map network takes the intermediate feature maps as input, and yields the content-weighted importance map $P(\mathbf{x})$.
The rounding function is adopted to quantize $P(\mathbf{x})$ and generate a mask $M(P(\mathbf{x}))$ that has the same size of $B(E(\mathbf{x}))$.
The binary code is then trimmed based on $M(P(\mathbf{x}))$. And the convolutional decoder defines a nonlinear synthesis transform to produce decoding result $\hat{\mathbf{x}}$.

In the following, we first introduce the components of the framework and then present the formulation and learning method of our model.

\subsection{Components and Gradient Computation}\label{sec3_1}

\subsubsection{Convolutional encoder and decoder}

Both the encoder and decoder in our framework are fully convolution networks and can be trained by back-propagation.
The encoder network consists of three convolutional layers and three residual blocks.
Following~\cite{he2016deep}, each residual block has two convolutional layers. We further remove the batch normalization operations from the residual blocks.
The input image $\mathbf{x}$ is first convolved with 128 filters with size $8 \times 8$ and stride 4 and followed by one residual block.
The feature maps are then convolved with 256 filters with size $4 \times 4$ and stride 2 and followed by two residual blocks to output the intermediate feature maps $f(\mathbf{x})$.
Finally, $f(\mathbf{x})$ is convolved with $m$ filters with size $1 \times 1$ to yield the encoder output $E(\mathbf{x})$.
It should be noted that we set $n = 64$ for low comparison rate models with less than $0.5$ bpp, and $n = 128$ otherwise.


The network architecture of decoder $D(\mathbf{c})$ is symmetric to that of the encoder, where $\mathbf{c}$ is the code of an image $\mathbf{x}$.
To upsample the feature maps, we adopt the depth to space operation mentioned in~\cite{toderici2016full}.
Please refer to the supplementary material for more details on the network architecture of the encoder and decoder.


\subsubsection{Binarizer}

Due to sigmoid nonlinearity is adopted in the last convolutional layer, the encoder output $\mathbf{e} = E(\mathbf{x})$ should be in the range of $[0, 1]$.
Denote by $e_{ijk}$ an element in $\mathbf{e}$.
The binarizer is defined as
\begin{equation}
B(e_{ijk})=\begin{cases}
1, & \mbox{if } e_{ijk} > 0.5\\
0, & \mbox{if } e_{ijk} \leq 0.5
\end{cases}
\end{equation}
%
%
However, the gradient of the binarizer function $B(e_{ijk})$ is zero almost everywhere except that it is infinite when $e_{ijk} = 0.5$.
In the back-propagation algorithm, the gradient is computed layer by layer by using the chain rule in a backward manner.
Thus, this will make any layer before the binarizer (\ie, the whole encoder) never be updated during training.


Fortunately, some recent works on binarized neural networks (BNN)~\cite{zhou2016dorefa,rastegari2016xnor,courbariaux2016binarized} have studied the issue of propagating gradient through binarization.
Based on the straight-through estimator on gradient~\cite{courbariaux2016binarized}, we introduce a proxy function $\tilde{B}(e_{ijk})$ to approximate $B(e_{ijk})$.
Here, $B(e_{ijk})$ is still used in forward propagation calculation, while $\tilde{B}(e_{ijk})$ is used in back-propagation.
Inspired by BNN, we adopt a piecewise linear function $\tilde{B}(e_{ijk})$ as the proxy of $B(e_{ijk})$,
\begin{equation}
\tilde{B}(e_{ijk})=\begin{cases}
1, & \mbox{if } e_{ijk} > 1\\
e_{ijk}, & \mbox{if }  1\leq e_{ijk} \leq 0 \\
0, & \mbox{if }  e_{ijk} < 0
\end{cases}
\end{equation}
Then, the gradient of $\tilde{B}(e_{ijk})$ can be easily obtained by,
\begin{equation}
\tilde{B}^\prime(e_{ijk})=\begin{cases}
1, & \mbox{if } 1\leq e_{ijk}\leq 0 \\
0, & \mbox{otherwise}
\end{cases}
\end{equation}

\subsubsection{Importance map} \label{sec3_1_3}

\begin{figure}
\begin{center}
\includegraphics[width=0.8\linewidth]{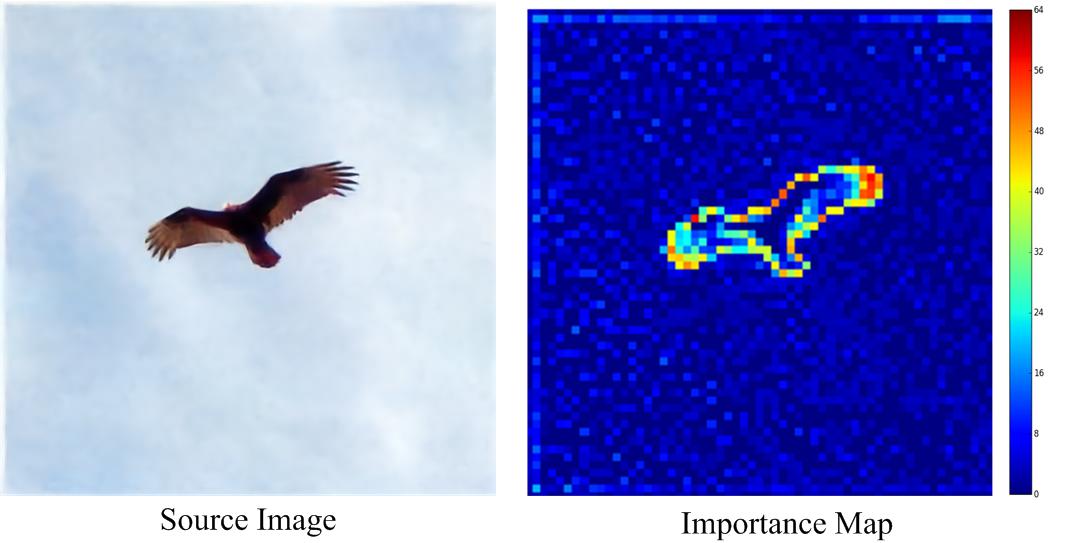}
\end{center}
   \caption{Illustration of importance map. The regions with sharp edge or rich texture generally have higher values and should be allocated more bits to encode.}
\label{importance_map}
\end{figure}

In~\cite{balle2016end,theis2017lossy}, the code length after quantization is spatially invariant, and entropy coding is then used to further compression the code.
Actually, the difficulty in compressing different parts of an image should be different.
The smooth regions in an image should be easier to be compressed than those with salient objects or rich textures.
Thus, fewer bits should be allocated to the smooth regions while more bits should be allocated to the regions with more information content.
For example, given an image with an eagle flying in the blue sky in Figure~\ref{importance_map}, it is reasonable to allocate more bits to the eagle and fewer bits to blue sky.
Moreover, when the whole code length for an image is limited, such allocation scheme can also be used for rate control.

We introduce a content-weighted importance map for bit allocation and compression rate control.
It is a feature map with only one channel, and its size should be same with that of the encoder output.
The value of importance map is in the range of $(0, 1)$.
An importance map network is deployed to learn the importance map from an input image $\mathbf{x}$.
It takes the intermediate feature maps $f(\mathbf{x})$ from the last residual block of the encoder as input, and uses a network of three convolutional layers to produce the importance map $\mathbf{p} = P({\mathbf{x}})$.

Denote by $h \times w$ the size of the importance map $\mathbf{p}$, and $n$ the number of feature maps of the encoder output.
In order to guide the bit allocation, we should first quantize each element in $\mathbf{p}$ to an integer no more than $n$, and then generate an importance mask $\mathbf{m}$ with size $ n \times h \times w$. 
Given an element $p_{ij}$ in $\mathbf{p}$, the quantizer to importance map is defined as,
\begin{equation}
Q(p_{ij}) = \begin{cases}

l-1, & \mbox{if } \frac{l-1}{L} \leq p_{ij} < \frac{l}{L}, l = 1, \ldots, L\\
L, & \mbox{if } p_{ij} = 1
\end{cases}
\end{equation}
where L is the importance levels and $(n\;mod\;L)=0$. Each important level is corresponding to $\frac{n}{L}$ bits.
As mentioned above, $p_{ij}\in (0,1)$. Thus, $p_{ij}\neq 1$ and $Q(p_{ij})$ has only L kinds of different quantity value i.e. $0,\ldots, L-1$.

It should be noted that, $Q(p_{ij}) = 0$ indicates that zero bit will be allocated to this location, and all its information can be reconstructed based on its context in the decoding stage.
By this way, the importance map can not only be treated as an alternative of entropy rate estimation but also naturally take the context into account.

With $Q(\mathbf{p})$, the importance mask $\mathbf{m} = M(\mathbf{p})$ can then be obtained by,
\begin{equation}
\mathbf{m}_{kij} = \begin{cases}
1, & \mbox{if } k \leq \frac{n}{L} Q(p_{ij})\\
0, & \mbox{else}
\end{cases}
\end{equation}
The final coding result of the image $\mathbf{x}$ can then be represented as $\mathbf{c} = M(\mathbf{p}) \circ B(\mathbf{e})$, where $\circ$ denotes the element-wise multiplication operation.
Note that the quantized importance map $Q(\mathbf{p})$ should also be considered in the code. Thus all the bits with $\mathbf{m}_{kij} = 0$ can be safely excluded from $B(\mathbf{e})$.
Thus, instead of $n$, only $\frac{n}{L}Q(p_{ij})$ bits are need for each location $(i, j)$.
Besides, in video coding, just noticeable distortion (JND) models~\cite{yang2005just} have also been suggested for spatially variant bit allocation and rate control.
Different from~\cite{yang2005just}, the importance map are learned from training data by optimizing joint rate-distortion performance.

Finally, in back-propagation, the gradient $\mathbf{m}$ with respect to ${p_{ij}}$ should be computed.
Unfortunately, due to the quantization operation and mask function, the gradient is zero almost everywhere.
Actually, the importance map $m$ can be equivalently rewritten as a function of $p$,
\begin{equation}\label{eq6}
\mathbf{m}_{kij} = \begin{cases}
1, & \mbox{if } \ceil{\frac{kL}{n}} < L p_{ij}+1\\
0, & \mbox{else}
\end{cases}
\end{equation}
where $\ceil{.}$ is the ceiling function.
Analogous to binarizer, we also adopt a straight-through estimator of the gradient,

\begin{equation}
\frac{\partial \mathbf{m}_{kij} }{\partial p_{ij}} = \begin{cases}
L,\;\mbox{if } Lp_{ij}-1\leq \ceil{\frac{kL}{n}}<Lp_{ij}+2 \\
0,\; \mbox{else}
\end{cases}
\end{equation}
\subsection{Model formulation and learning}

\subsubsection{Model formulation}

In general, the proposed content-weighted image compression system can be formulated as a rate-distortion optimization problem.
Our objective is to minimize the combination of the distortion loss and rate loss.
A tradeoff parameter $\gamma$ is introduced for balancing compression rate and distortion.
Let $\mathcal{X}$ be a set of training data, and $\mathbf{x} \in \mathcal{X}$ be an image from the set.
Therefore, the objective function our model is defined as
\begin{equation}
\mathcal{L} =\sum_{\mathbf{x} \in \mathcal{X}}\{\mathcal{L}_{D}(\mathbf{c}, \mathbf{x}) + \gamma \mathcal{L}_{R}(\mathbf{x})\},
\end{equation}
where $\mathbf{c}$ is the code of the input image $\mathbf{x}$.
$\mathcal{L}_{D}(\mathbf{c}, \mathbf{x})$ denotes the distortion loss and $\mathcal{L}_{R}(\mathbf{x})$ denotes the rate loss, which will be further explained in the following.

\textbf{Distortion loss.} Distortion loss is used to evaluate the distortion between the original image and the decoding result.
Even better results may be obtained by assessing the distortion in the perceptual space.
With the input image $\mathbf{x}$ and decoding result $D(\mathbf{c})$, we simply use the squared $\ell_2$ error to define the distortion loss,
\begin{equation}
\mathcal{L}_{D}(\mathbf{c}, \mathbf{x}) = \|D(\mathbf{c}) - \mathbf{x}\|^2_2.
\end{equation}

\textbf{Rate loss.} Instead of entropy rate, we define the rate loss directly on the continuous approximation of the code length.
Suppose the size of encoder output $E(\mathbf{x})$ is $n \times h \times w$.
The code by our model includes two parts:
(i) the quantized importance map $Q(\mathbf{p})$ with the fixed size $h \times w$;
(ii) the trimmed binary code with the size $\frac{n}{L}\sum_{i,j} Q({p}_{ij})$.
Note that the size of $Q(\mathbf{p})$ is constant to the encoder and importance map network.
Thus $\frac{n}{L}\sum_{i,j} Q({p}_{ij})$ can be used as rate loss.

Due to the effect of quantization $Q({p}_{ij})$, the function $\frac{n}{L}\sum_{i,j} Q({p}_{ij})$ cannot be optimized by back-propagation.
Thus, we relax $Q({p}_{ij})$ to its continuous form, and use the sum of the importance map $\mathbf{p} = P(\mathbf{x})$ as rate loss,
%
%
%
%
\begin{equation}
\label{eq:rateloss0}
\mathcal{L}_{R}^0(\mathbf{x}) = \sum_{i,j} (P(\mathbf{x}))_{ij}
\end{equation}
For better rate control, we can select a threshold $r$, and penalize the rate loss in Eqn. (\ref{eq:rateloss0}) only when it is higher than $r$.
And we then define the rate loss in our model as,
%
\begin{equation}
\mathcal{L}_{R}(\mathbf{x}) \!=\! \begin{cases}
 \sum_{i,j} (P(\mathbf{x}))_{ij} \!-\! r, \!&\! \mbox{if }  \sum_{i,j} (P(\mathbf{x}))_{ij} \!>\! r\\
 0, \!&\! \mbox{otherwise}
 \end{cases}
\end{equation}
The threshold $r$ can be set based on the code length for a given compression rate.
By this way, our rate loss will penalize the code length higher than $r$, and makes the learned compression system achieve the comparable compression rate to the given one.


\subsubsection{Learning}

Benefited from the relaxed rate loss and the straight-through estimator of the gradient, the whole compression system can be trained in an end-to-end manner with an ADAM solver~\cite{kingma2014adam}.
We initialize the model with the parameters pre-trained on training set $\mathcal{X}$ without the importance map.
The model is further trained with the learning rate of $1e^{-4}$, $1e^{-5}$ and $1e^{-6}$.
In each learning rate, the model is trained until the objective function does not decrease.
And a smaller learning rate is adopted to fine-tune the model.

\section{Convolutional entropy encoder}\label{sec4}

Due to no entropy constraint is included, the code generated by the compression system in Sec.~\ref{sec3} is non-optimal in terms of entropy rate.
This provides some leeway to further compress the code with lossless entropy coding.
Generally, there are two kinds of entropy compression methods, \ie Huffman tree and arithmetic coding~\cite{witten1987arithmetic}.
Among them, arithmetic coding can exhibit better compression rate with a well-defined context, and is adopted in this work.

\subsection{Encoding binary code}

The binary arithmetic coding is applied according to the CABAC~\cite{marpe2003context} framework.
Note that CABAC is originally proposed for video compression.
Let $\mathbf{c}$ be the code of $n$ binary bitmaps, and $\mathbf{m}$ be the corresponding importance mask.
To encode $\mathbf{c}$, we modify the coding schedule, redefine the context, and use CNN for probability prediction.
As to coding schedule, we simply code each binary bit map from left to right and row by row, and skip those bits with the corresponding important mask value of 0.


\begin{figure}
\begin{center}
\includegraphics[width=1\linewidth]{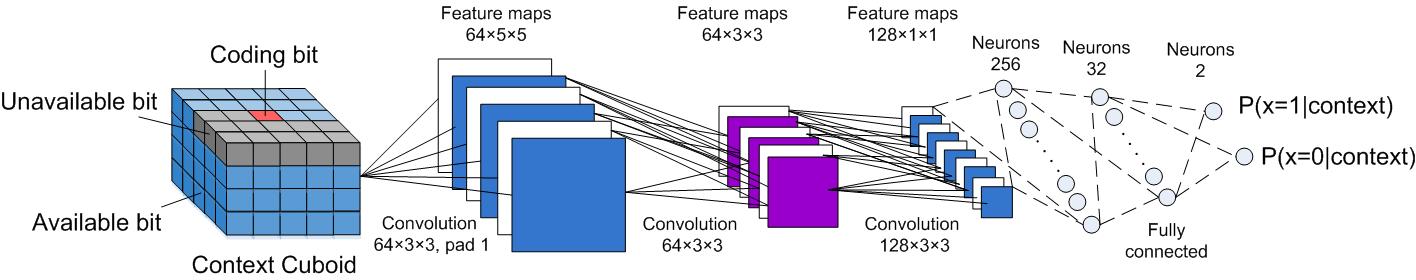}
\end{center}
   \caption{The CNN for convolutional entropy encoder. The red block represents the bit to predict; dark blocks mean unavailable bits; blue blocks represent available bits.}
\label{en_encoder}
\end{figure}

\textbf {Context modeling.} Denote by $c_{kij}$ be a binary bit of the code $\mathbf{c}$.
We define the context of $c_{kij}$ as $CNTX(c_{kij})$ by considering the binary bits both from its neighbourhood and from the neighboring maps.
%
%
Specifically, $CNTX(c_{kij})$ is a $5\times 5 \times 4$ cuboid.
We further divide the bits in $CNTX(c_{kij})$ into two groups: the available and unavailable ones.
The available ones represent those can be used to predict $c_{kij}$.
While the unavailable ones include: (i) the bit to be predicted $c_{kij}$, (ii) the bits with the importance map value 0, (iii) the bits out of boundary and (iv) the bits currently not coded due to the coding order.
Here we redefine $CNTX(c_{kij})$ by: (1) assigning 0 to the unavailable bits, (2) assigning 1 to the unavailable bits with value 0, and assigning 2 to the unavailable bits with value 1.
%

\textbf{Probability prediction.} One usual method for probability prediction is to build and maintain a frequency table.
As to our task, the size of the cuboid is too large to build the frequency table.
Instead, we introduce a CNN model for probability prediction.
As shown in Figure~\ref{en_encoder}, the convolutional entropy encoder $En(CNTX(c_{kij}))$ takes the cuboid as input, and output the probability that the bit $c_{kij}$ is 1.
Thus, the loss for learning the entropy encoder can be written as,
\begin{align}
&\mathcal{L}_{E} = \sum_{i,j,k}
 m_{kij} \left\{ c_{kij}\log_2(En(CNTX(c_{kij})))\nonumber \right.\\
& \left. +(1-c_{kij})\log_2(1-En(CNTX(c_{kij}))) \right\}
\end{align}
where $\mathbf{m}$ is the importance mask. 
The convolutional entropy encoder is trained using the ADAM solver on the contexts of binary codes extracted from the binary feature maps generated by the trained encoder.
The learning rate decreases from $1e^{-4}$ to $1e^{-6}$ as we do in Sec. \ref{sec3}.


\subsection{Encoding quantized importance map}

We also extend the convolutional entropy encoder to the quantized importance map.
To utilize binary arithmetic coding, a number of binary code maps are adopted to represent the quantized importance map.
%
%
The convolutional entropy encoder is then trained to compress the binary code maps.


\section{Experiments}
\begin{figure}
\centering
\label{mse} 
\includegraphics[width=1\linewidth]{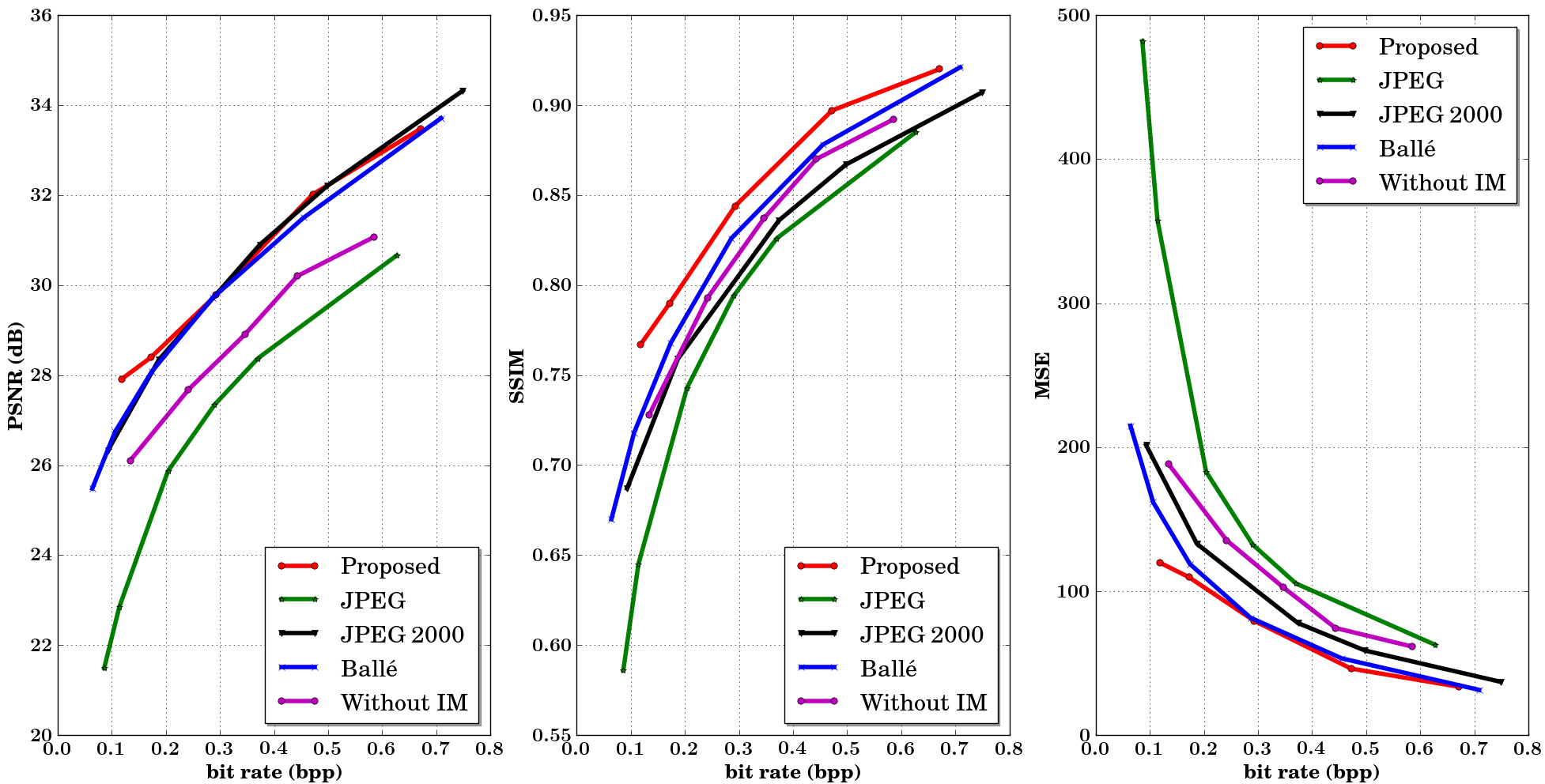}

\caption{Comparison of the ratio-distortion curves by different methods: (a) PSNR, (b) SSIM, and (c) MSE.\quad "Without IM" represents the proposed method without importance map.}
\label{ratio_distortion} 
\end{figure}

\begin{figure*}
\begin{center}
\subfigure{
\includegraphics[width=1\linewidth]{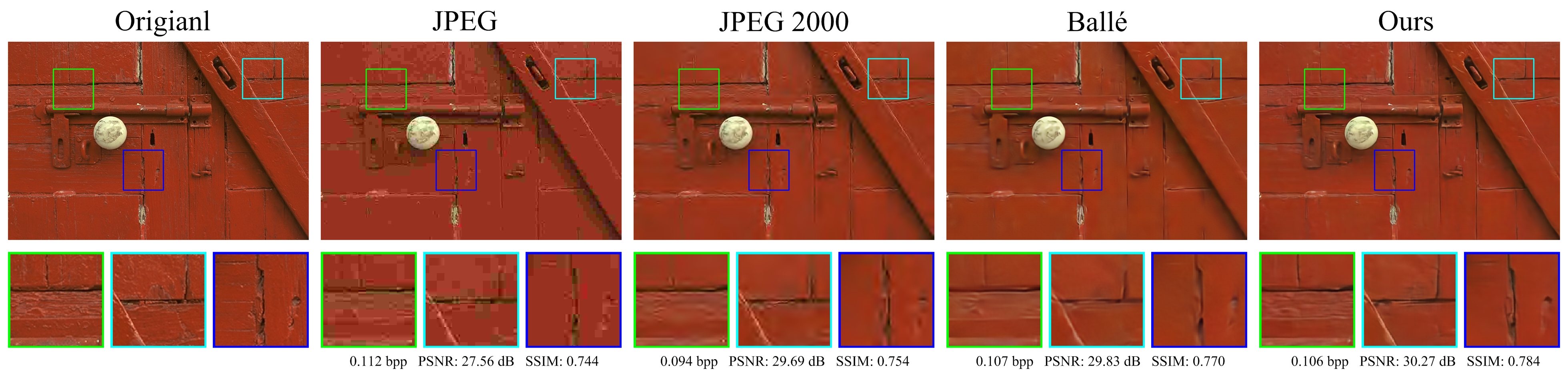}
}
\subfigure{
\includegraphics[width=1\linewidth]{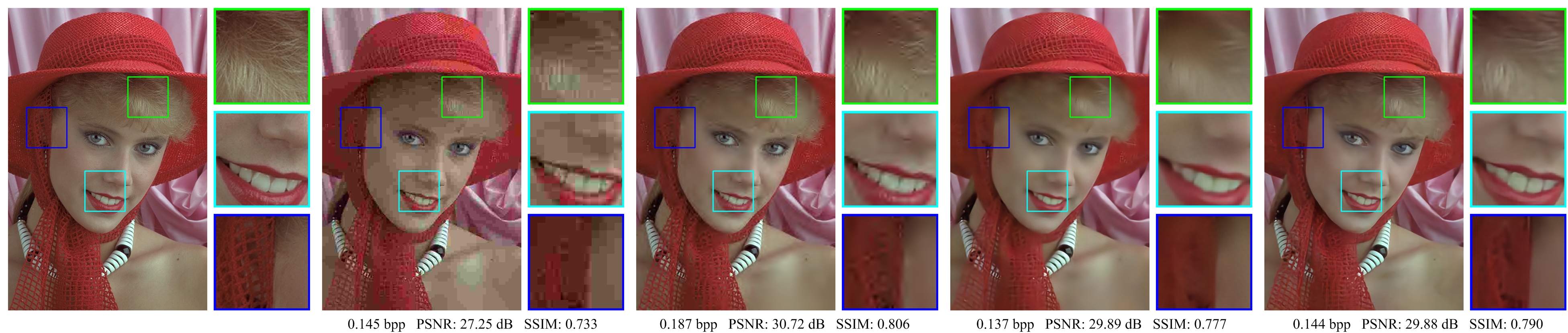}
}
\subfigure{
\includegraphics[width=1\linewidth]{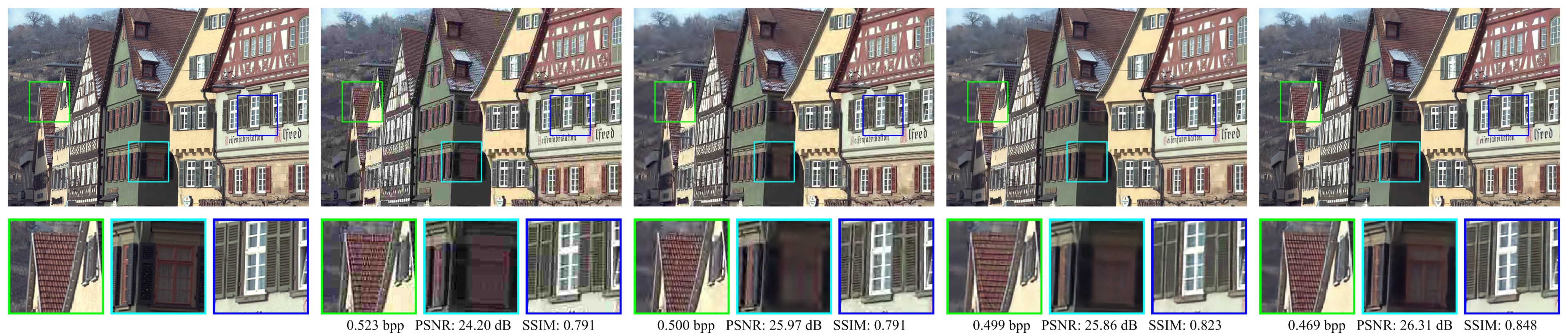}}
\subfigure{
\includegraphics[width=1\linewidth]{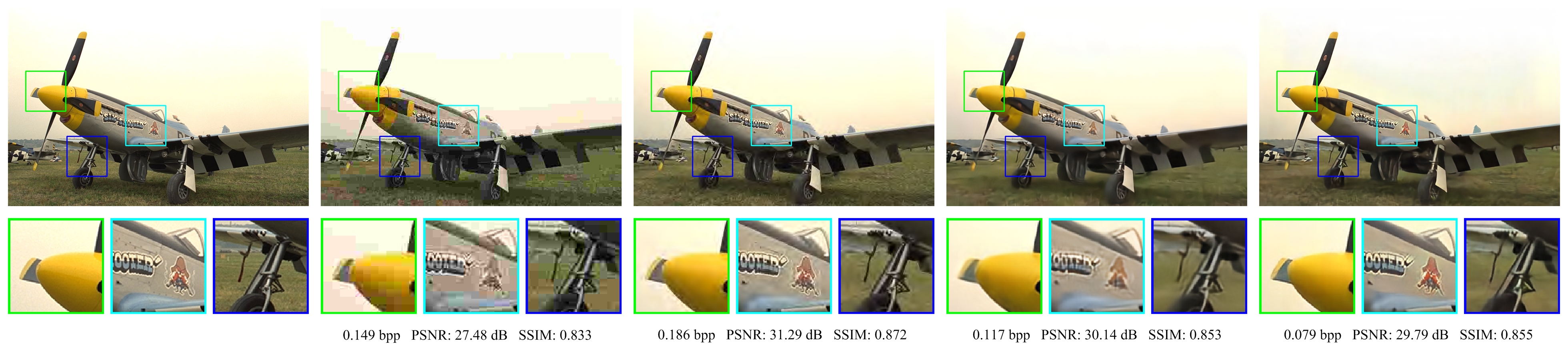}}
\subfigure{
\includegraphics[width=1\linewidth]{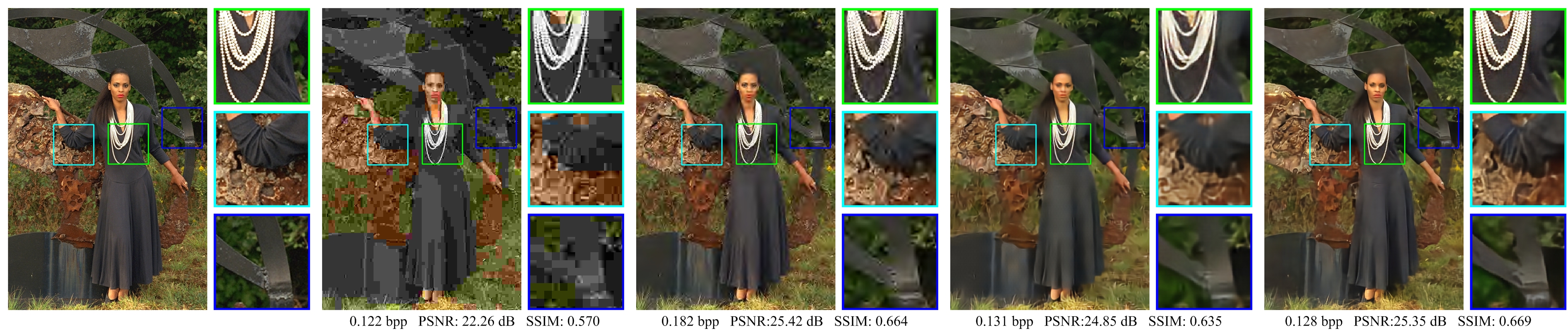}}
\end{center}
   \caption{Images produced by different compression systems at different compression rates. From the left to right: groundtruth, JPEG, JPEG 2000, Ball{\'e}~\cite{balle2016end}, and ours. Our model achieves the best visual quality at each rate, demonstrating the superiority of our model in preserving both sharp edges and detailed textures. (Best viewed on screen in color)}
\label{visual_results}
\end{figure*}
Our content-weighted image compression model are trained on a subset of ImageNet~\cite{deng2009imagenet} with about $10,000$ high quality images. 
We crop these images into $128\times 128$ patches and take use of these patches to train the network. 
After training, we test our model on the Kodak PhotoCD image dataset with the metrics for lossy image compression.
The compression rate of our model is evaluated by the metric bits per pixel (bpp), which is calculated as the total amount of bits used to code the image divided by the number of pixels.
The image distortion is evaluated with Means Square Error (MSE), Peak Signal-to-Noise Ratio (PSNR), and the structural similarity (SSIM) index.

In the following, we first introduce the parameter setting of our compression system. 
Then both quantitative metrics and visual quality evaluation are provided. 
Finally, we further analyze the effect of importance map and convolutional entropy encoder on the compression system.

\subsection{Parameter setting}


In our experiments, we set the number of binary feature maps $n$ according to the compression rate, \ie 64 when the compression rate is less than $0.5$ bpp and 128 otherwise.
Then, the number of importance level is chosen based on $m$. 
For $n = 64$ and $n = 128$, we set the number of importance level to be $16$ and $32$, respectively.
Moreover, different values of the tradeoff parameter $\gamma$ in the range $[0.0001, 0.2]$ are chosen to get different compression rates. 
%
For the choice of the threshold value $r$, we just set it as $r_0 h w$ for $n=64$ and $0.5r_0 h w$ for $n=128$. $r_0$ is the wanted compression rate represent with bit per pixel (bpp).

\subsection{Quantitative evaluation}
For quantitative evaluation, we compare our model with JPEG~\cite{wallace1992jpeg}, JPEG 2000~\cite{skodras2001jpeg}, and the CNN-based method by Ball{\'e} \etal~\cite{balle2016end}.
Among the different variants of JPEG, the optimized JPEG with 4:2:0 chroma sub-sampling is adopted. 
For the sake of fairness, all the results by Ball{\'e}~\cite{balle2016end}, JPEG, and JPEG2000 on the Kodak dataset are downloaded from \url{ http://www.cns.nyu.edu/~lcv/iclr2017/}.
%

Using MSE, SSIM~\cite{wang2004image} and PSNR as performance metrics, Figure~\ref{ratio_distortion} gives the ratio-distortion curves of these four methods. 
In terms of MSE, JPEG has the worst performance. 
And both our system and Ball{\'e}~\cite{balle2016end} can be slightly better than JPEG 2000.
In terms of PSNR, the results by JPEG 2000, Ball{\'e}~\cite{balle2016end} and ours are very similar, but are much higher than that by JPEG.
In terms of SSIM, our system outperforms all the three competing methods, including JPEG, JPEG 2000, and Ball{\'e}~\cite{balle2016end}.
Due to SSIM is more consistent with human visual perception than PSNR and MSE, these results indicate that our system may perform better in terms of visual quality.
%


\subsection{Visual quality evaluation}

We further compare the visual quality of the results by JPEG, JPEG 2000, Ball{\'e}~\cite{balle2016end} and our system in low compression rate setting. 
Figure~\ref{visual_results} shows the original images and the results produced by the four compression systems.
Visual artifacts, \eg, blurring, ringing, and blocking, usually are inevitable in the compressed images by traditional image compression standards such as JPEG and JPEG 2000.
And these artifacts can also be perceived in the second and third columns of Figure~\ref{visual_results}.
Even Ball{\'e}~\cite{balle2016end} is effective in suppressing these visual artifacts.
In Figure~\ref{visual_results}, from the results produced by Ball{\'e}~\cite{balle2016end}, we can observe the blurring artifacts in row 1, 2, 3, and 5, the color distortion in row 4 and 5, and the ringing artifacts in row 4 and 5.
In contrast, the results produced by our system exhibit much less noticeable artifacts and are visually much more pleasing.

From Figure~\ref{visual_results}, Ball{\'e}~\cite{balle2016end} usually produces the results by blurring the strong edges or over-smoothing the small-scale textures.
Specifically, in row 5 most details of the necklace have been removed by Ball{\'e}~\cite{balle2016end}.
One possible explanation may be that before entropy encoding it adopts a spatially invariant bit allocation scheme. 
Actually, it is natural to see that more bits should be allocated to the regions with strong edges or detailed textures while less to the smooth regions.
Instead, in our system, an importance map is introduced to guide spatially variant bit allocation.
Moreover, instead of handcrafted engineering, the importance map are end-to-end learned to minimize the rate-distortion loss. 
As a result, our model is very promising in keeping perceptual structures, such as sharp edges and detailed textures.

\subsection{Experimental analyses on important map}

\begin{figure}
\begin{center}
\includegraphics[width=0.9\linewidth]{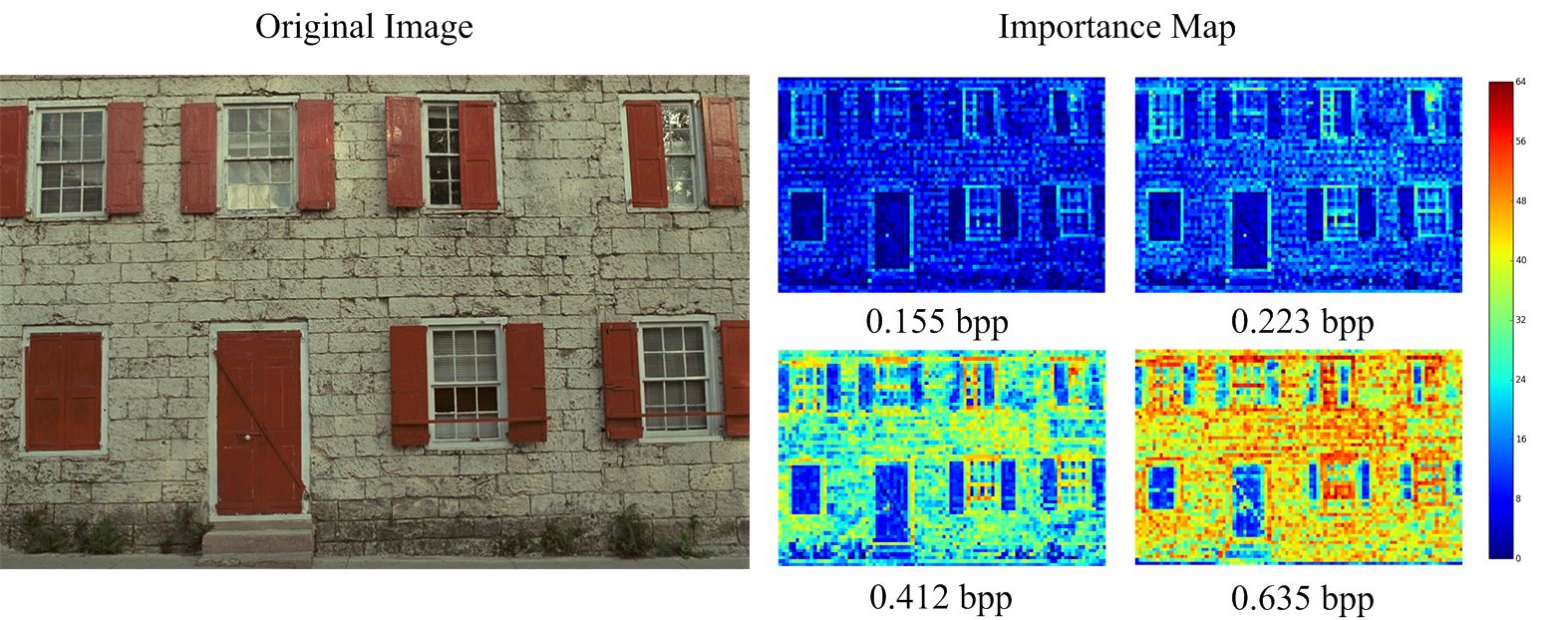}
\end{center}
   \caption{The important maps obtained at different compression rates. The right color bar shows the palette on the number of bits.}
\label{imp_map}
\end{figure}

To assess the role of importance map, we train a baseline model by removing the importance map network from our framework.
Both entropy and importance map based rate loss are not included in the baseline model.
Thus, the compression rate is controlled by modifying the number of binary feature maps.
Figure~\ref{ratio_distortion} also provides the ratio-distortion curves of the baseline model.
%
%
One can see that, the baseline model performs poorer than JPEG 2000 and Ball{\'e}~\cite{balle2016end} in terms of MSE, PSNR, and SSIM,
validating the necessity of importance map for our model. 
Using the image in row 5 of Figure~\ref{visual_results}, the compressed images by our model with and without importance map are also shown in the supplementary material.
And more detailed textures and better visual quality can be obtained by using the importance map.


Figure~\ref{imp_map} shows the importance map obtained at different compression rates. 
%
%
One can see that, when the compression rate is low, due to the overall bit length is very limited, the importance map only allocates more bits to salient edges.
With the increasing of compression rate, more bits will begin to be allocated to weak edges and mid-scale textures.
Finally, when the compression rate is high, small-scale textures will also be allocated with more bits.  
Thus, the importance map learned in our system is consistent with human visual perception, which may also explain the advantages of our model in preserving the structure, edges and textures.
%

\subsection{Entropy encoder evaluation}
\begin{figure}
\centering
\includegraphics[width=0.75\linewidth]{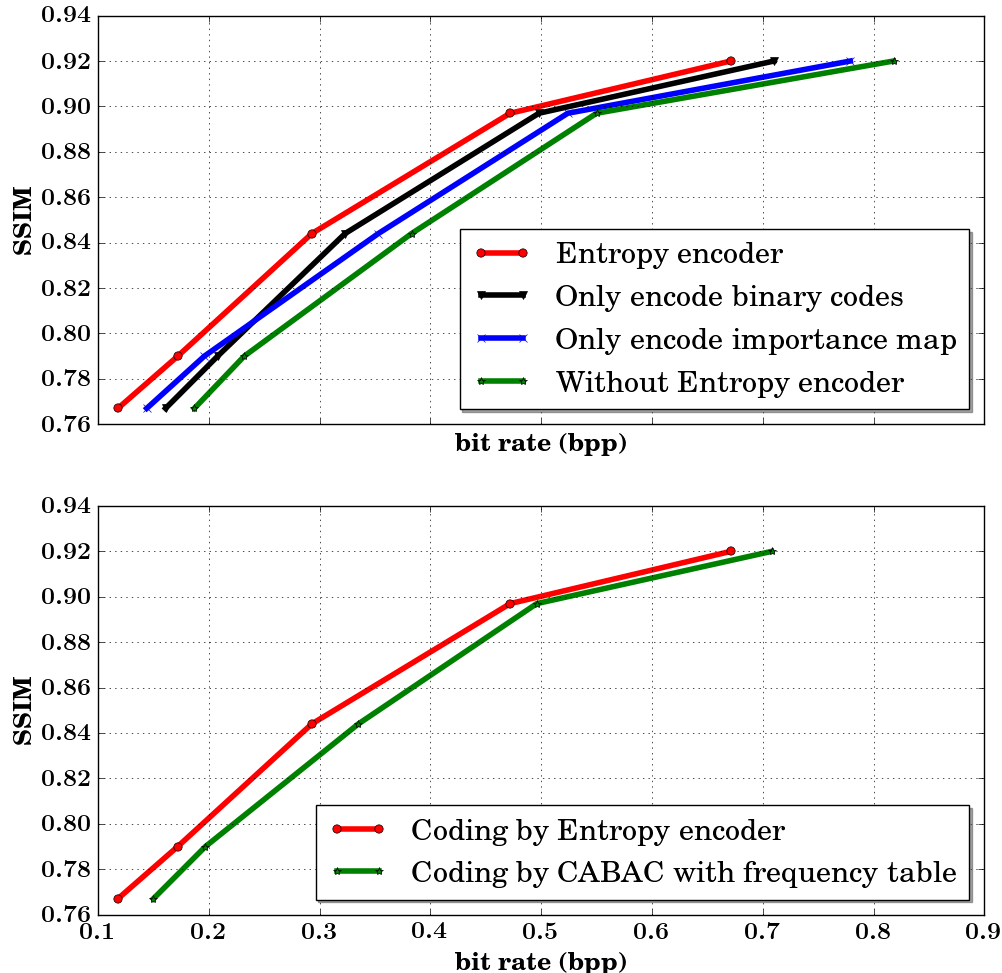}
\caption{Performance of convolutional entropy encoder: (a) for encoding binary codes and importance map, and (b) by comparing with CABAC. }
\label{cmp_abac} 
\end{figure}

%
The model in Sec. \ref{sec3} does not consider entropy rate, allowing us to further compress the code with convolutional entropy encoder. 
Here, two groups of experiments are conducted.
First, we compare four variants of our model: (i) the full model, (ii) the model without entropy coding, (iii) the model by only encoding binary codes, and (iv) the model by only encoding importance map.
From Figure~\ref{cmp_abac}{\color{red}(a)}, both the binary codes and importance map can be further compressed by using our convolutional entropy encoder.
And our full model can achieve the best performance among the four variants.
Second, we compare our convolutional entropy coding with the standard CABAC with small context (\ie the 5 bits near the bit to encode). 
As shown in Figure~\ref{cmp_abac}{\color{red}(b)}, our convolutional entropy encoder can take larger context into account and perform better than CABAC.
Besides, we also note that our method with either CABAC or convolutional encoder can outperform JPEG 2000 in terms of SSIM.


\section{Conclusion}

A CNN-based system is developed for content weighted image compression. 
With the importance map, we suggest a non-entropy based loss for rate control.
Spatially variant bit allocation is also allowed to emphasize the salient regions. 
Using the straight-through estimator, our model can be end-to-end learned on a training set.
A convolutional entropy encoder is introduced to further compress the binary codes and the importance map. 
Experiments clearly show the superiority of our model in retaining structures and removing artifacts, leading to remarkable visual quality.

{\small
\bibliographystyle{ieee}
\bibliography{egbib}
}
\clearpage
\appendix

\section{Network Architecture}
\begin{table}[htb]
\scriptsize
\begin{center}
\begin{tabular}{c|c}
\hline
Layer & Activation size \\
\hline
Input & $3\times 128\times 128$ \\
$8\times8\times128$ conv, pad $2$, stride $4$ & $128\times32\times32$ \\
Residual block, 128 filters & $128\times32\times32$ \\
$4\times4\times256$ conv, pad $1$, stride $2$ & $256\times16\times16$ \\
Residual block, 256 filters & $256\times16\times16$ \\
Residual block, 256 filters & $256\times16\times16$ \\
$1\times1\times64(128)$ conv, pad $0$, stride $1$ & $64(128)\times16\times16$ \\
\hline
\end{tabular}
\end{center}
\caption{Network architecture of the convolutional encoder.}
\label{table:encoder}
\end{table}

\begin{table}[htb]
\scriptsize
\begin{center}
\begin{tabular}{c|c}
\hline
Layer & Activation size \\
\hline
Input & $64(128)\times 16\times 16$ \\
$1\times1\times512$ conv, pad $0$, stride $1$ & $512\times16\times16$ \\
Residual block, 512 filters & $512\times16\times16$ \\
Residual block, 512 filters & $512\times16\times16$ \\
Depth to Space, stride 2 & $128\times32\times32$ \\
$3\times3\times256$ conv, pad $1$, stride $1$ & $256\times32\times32$ \\
Residual block, 256 filters & $256\times32\times32$ \\
Depth to Space, stride 4 & $128\times32\times32$ \\
$3\times3\times32$ conv, pad $1$, stride $1$ & $32\times128\times128$ \\
$3\times3\times3$ conv, pad $1$, stride $1$ & $3\times128\times128$ \\
\hline
\end{tabular}
\end{center}
\caption{Network architecture of the convolutional encoder.}
\label{table:decoder}
\end{table}

\begin{figure}
\begin{center}
\includegraphics[width=0.6\linewidth]{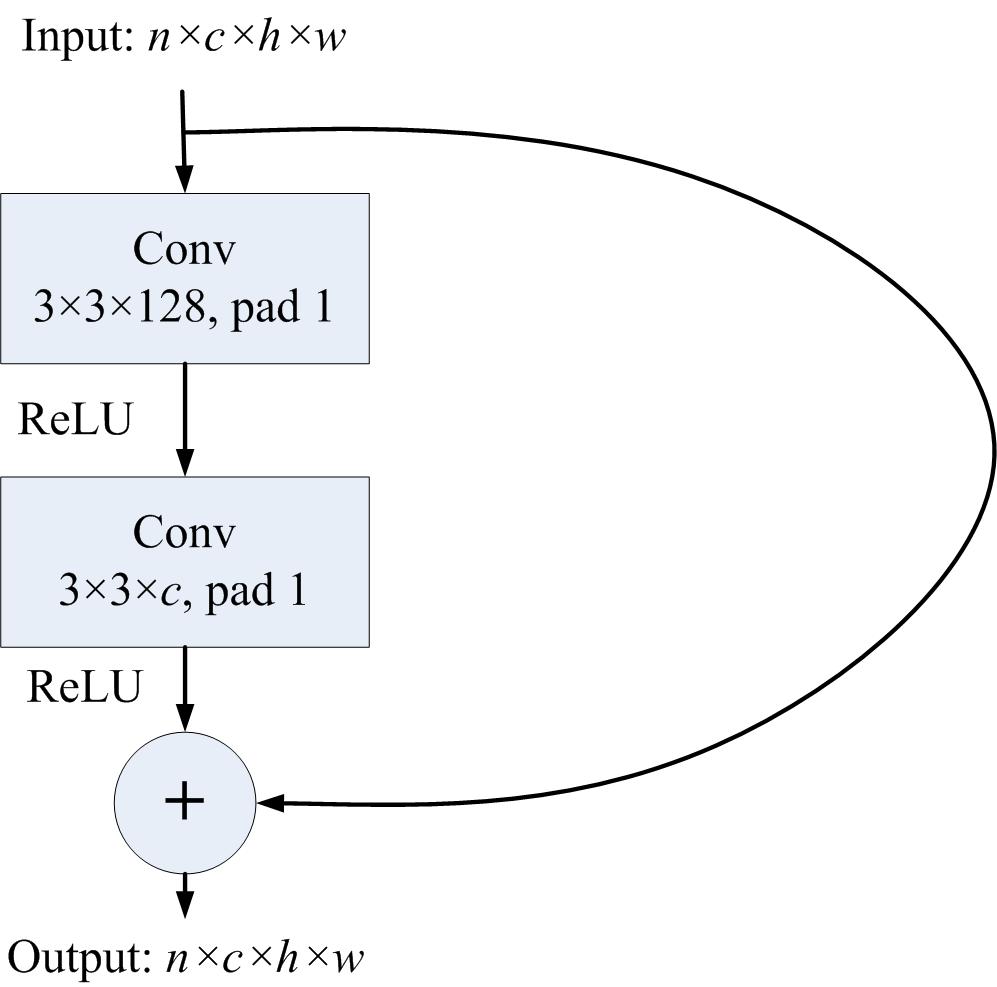}
\end{center}
   \caption{Structure of the residual blocks.}
\label{residue}
\end{figure}

Table~\ref{table:encoder} and Table~\ref{table:decoder} give the network architectures of the convolutional encoder and decoder, respectively. Except for the last layer, each convolutional layer is followed by ReLU nonlinearity. For the encoder, the last convolutional layer is followed by a Sigmoid nonlinearity to make sure the output of the encoder is the interval of $(0,1)$.  As to the decoder, there is no nonlinear layer after the last convolutional layer. For the residual block, we stack two convolutional layers in each block and remove the batch normalization layers. The architecture of the residual blocks is shown in Figure~\ref{residue}.

\section{Binarizing scheme for importance map}
The importance map is a part of our codes. In order to compress the importance map with binary arithmetic coding method, we should first binarize the importance map. In this work, we simply adopt the binary representation of the quantized importance map $Q(\mathbf{p})$ to generate the binary importance map
$\mathbf{b}=B(Q(\mathbf{p}))$ with the shape of $n_b\times w \times h$. Here, $n_b$ is the number of feature maps in the binary importance map and it satisfies $2^{n_b-1} < L \leq 2^{n_b}$. $L$ is the importance levels. Given the importance map $Q(\mathbf{p})$, the binary importance map $\mathbf{b}$ follows the equation below.
\begin{equation}\label{eq1}
Q(\mathbf{p})=\sum_{k=0}^{n_b-1}\mathbf{b}_{kij}2^k
\end{equation}
With Eqn.~\ref{eq1}, the binary importance map can be easily calculated from the quantized importance map.
\section{Experiments supplementary}

\begin{figure*}
\begin{center}
\includegraphics[width=0.75\linewidth]{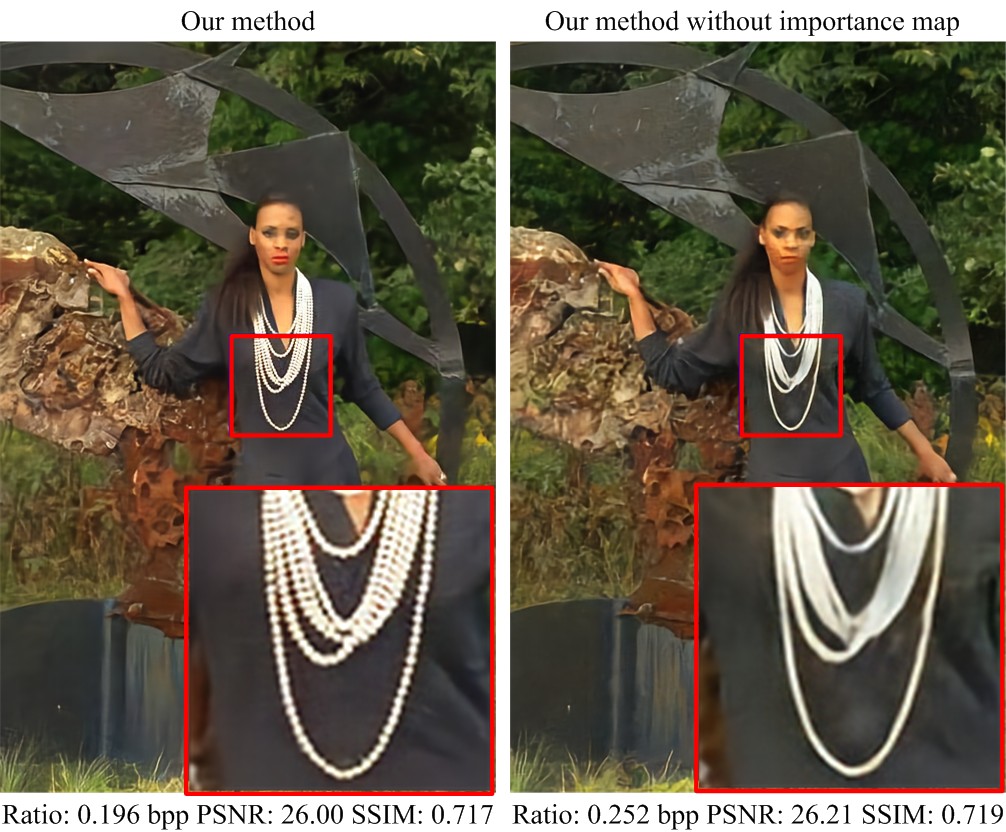}
\end{center}
   \caption{Comparison between our model with and without importance map.}
\label{cmp}
\end{figure*}

The compressed images by our model with and without importance map are also shown in Figure~\ref{cmp}. And more detailed textures and better visual quality can be obtained by using the importance map. This indicates that the introduced importance map provides our model with more ability to model the textures and edges in low bit rate image compression.

More high resolution results can be found at \url{http://www2.comp.polyu.edu.hk/~15903062r/content-weighted-image-compression.html}. And a large vision of our paper with more experiment results in the appendix is also available at this site.

\clearpage

\end{document}